\documentclass[UTF8]{article}
\usepackage[T1]{fontenc}
\usepackage[utf8]{inputenc}
\usepackage{authblk}
\usepackage{amsmath}
\usepackage{amsfonts}
\newtheorem{theorem}{Theorem}

\newtheorem{proof}{Proof}[section]
\usepackage{graphicx}
\usepackage{subfigure} 
\usepackage{float}
\usepackage{subfigure} 
\usepackage{algorithm}
\usepackage[noend]{algpseudocode}
\usepackage{tikz,xcolor,hyperref}
\definecolor{lime}{HTML}{A6CE39}
\DeclareRobustCommand{\orcidicon}{
	\begin{tikzpicture}
	\draw[lime, fill=lime] (0,0) 
	circle [radius=0.16] 
	node[white] {{\fontfamily{qag}\selectfont \tiny ID}};
	\draw[white, fill=white] (-0.0625,0.095) 
	circle [radius=0.007];
	\end{tikzpicture}
	\hspace{-2mm}
}
\foreach \x in {A, ..., Z}{\expandafter\xdef\csname orcid\x\endcsname{\noexpand\href{https://orcid.org/\csname orcidauthor\x\endcsname}
		{\noexpand\orcidicon}}
}

\usetikzlibrary{graphs}
\usetikzlibrary{shapes.geometric,fit,positioning,arrows,automata,calc}
\tikzset{
	main/.style={circle, minimum size = 3mm, thick, draw =black!80, node distance = 8mm},
	connect/.style={-latex, thick},
	box/.style={rectangle, draw=black!100}
}
\title{Decision Machines: Congruent Decision Trees}
\author{Jinxiong Zhang  \orcidA{}\\
jinxiongzhang@qq.com}
\date{}

\begin{document}

\maketitle

\begin{abstract}

    The decision tree recursively partitions the input space into regions 
    and derives axis-aligned decision boundaries from data.
    Despite its simplicity and interpretability, decision trees lack parameterized representation,
    which makes it prone to overfitting and difficult to find the optimal structure.
    We propose  Decision Machines, which embed Boolean tests into a binary vector space
    and represent the tree structure as a matrices,
    enabling an interleaved traversal of decision trees through matrix computation.
    Furthermore, we explore the congruence of decision trees and attention mechanisms, 
    opening new avenues for optimizing decision trees and potentially enhancing their predictive power.

\end{abstract}

\section{Introduction}

Decision trees have long been a cornerstone in the field of machine learning, 
revered for their simplicity and interpretability. 
However, their common induction methods, 
which are inherently greedy and data-driven, 
often lead to a representation that is less than optimal for computational efficiency. 
The training process recursively partitions the training space, 
dividing it into increasingly smaller subsets until a stopping criterion is met, 
resulting in a tree structure where leaf nodes predict based on the majority of their ground truth set. 
This approach, while effective, culminates in an implementation that is rule-based and less amenable to machine execution.

In contrast, other supervised learning methods have moved towards end-to-end paradigms, 
which offer more flexibility and efficiency. 
The quest for a more computationally friendly representation of decision trees has led to various innovations, 
including vectorized implementations and attempts to translate decision trees into neural networks. 
There are some end-to-end training schemes instead of the greedy training methods for decision trees
such as \cite{norouzi2015efficient, kontschieder2015deep, yang2018deep, hehn2019end, DCT}.
And  the implementation of decision trees can be vectorized
such as \cite{lucchese2015quickscorer, dato2016fast, lettich2016gpu, lucchese2017quickscorer} and RapidScorer\cite{ye2018rapidscorer}.
The oblique decision trees are induced from the derivatives of neural models \cite{lee2020oblique}. 	
In the section 2 of \cite{fridedman1991multivariate}, there is an equivalent model of recursive partitioning in the form of an
expansion in a set of basis functions (as products of different step functions)
by replacing the geometrical concepts of regions and splitting with the arithmetic notions of adding and multiplying.
And it is to transform each tree into a  polynomial form in \cite{kuralenok2019monoforest}.
There are also some schemes to translate the decision trees into neural networks such as \cite{EntropyNet, welbl2014casting, biau2019neural}.
This paper proposes Decision Machines (DMs), a paradigm that reframes binary decision trees in matrix computation, 
thereby improving both efficiency and interpretability.

DMs offer a compact representation of binary decision trees, 
explicitly drawing dependencies between predictions and Boolean tests. 
The binary decision tree is decomposed into two parts: the tree structure representation and the Boolean tests embedding. 
The binary tree structure is represented by a character matrix
and the evaluation of Boolean tests are embedded into a binary vector space. 
DMs navigate the input insatnce from the root to its destined exit leaf based on the mutiplication of the tree matrix and the embedding vector. 
This formulation not only enhances the interpretability of decision trees 
but also paves the way for leveraging deep learning toolkits to expedite their execution.

Furthermore, this paper explores the congruence of decision trees and attention mechanisms, 
a revolutionary concept in the realm of deep learning,  
in order to open up new avenues for optimizing decision trees and potentially enhance their predictive power. 
Our contributions include a provable algorithm for representing binary trees analytically, 
a novel interpretation of binary decision trees as error-correcting output codes, 
and some evidence supporting the derivation of attention mechanisms from decision trees.
     
\section{Logical Decision Machines}\label{sec2}

We represent binary decision trees in the language of matrix computation, mapping the true decision path as the maximum likelihood path. 
The logical decision machines embed Boolean tests into a binary space, 
enabling an interleaved traversal of decision trees through matrix computation.

\subsection{Construction}

We adopt the description of binary decision trees from \cite{lucchese2015quickscorer}.
\begin{quote}
Each decision tree $T = (\mathcal{N}, \mathcal{L})$ is composed of a set of internal nodes $\mathcal{N}$, and a set of leaves $\mathcal{L}$. 
Each $n \in \mathcal{N}$ is associated with a Boolean test over a specific feature.
Every leaf $l \in \mathcal{L}$ stores the prediction.
Considering the test result of a certain internal node $n \in \mathcal{N}$,  it is able to infer that some leaves cannot be the exit leaf.
Indeed, if $n$ is a false node (i.e., its test condition is false), the leaves in the left subtree of $n$ cannot be the exit leaf. 
Similarly, if $n$ is a true node, the leaves in the right subtree of $n$ cannot be the exit leaf. 
Once all the nodes have been processed, the only leaf left is exactly the exit leaf.
\end{quote}

For simplicity, we only consider the numerical variables to test, i.e., the input instance $\mathbf{x}$ is a real vector in $\mathbb{R}^n$.
And Boolean conditions are used to test 
whether the numerical variables are no greater than some given specific thresholds.

In order to represent the true conditions as $-1$s, we  substitute $0$ with $-1$ in the sign function when $0$.
And we define the $\operatorname{sgn}(z)$ for $z\in\mathbb{R}$ in our context by
  \begin{equation}\label{signum}
  \operatorname{sgn}(z)=\begin{cases}
  -1,  &\text{if $z\leq 0$},\\
  +1,  &\text{if $z>0$}.
  \end{cases}
  \end{equation}

We use the hidden layer to generate the results of all binary tests in decision tree defined by
\begin{equation}\label{result}
     h(\mathbf{x})=\operatorname{sgn}(\mathrm{S}\mathbf{x}-\mathbf{t}),
\end{equation}
the matrix $\mathrm{S}$ is of $\{0, 1\}^{(L-1)\times  n}$ acting as variable selection to test;
the instance $\mathbf{x}$ is a real vector in $\mathbb{R}^n$;
the parameter $\mathbf{t}$ is a real vector in $\mathbb{R}^{L-1}$;
the integer $L$ is the total number of the leaves.

\begin{figure}[htb] 
     \tikzstyle{results} = [circle, text centered, draw=black]
     \tikzstyle{decisions} = [rectangle, rounded corners, text centered, draw = gray]
     \tikzstyle{arrow} = [->, >=stealth, draw=blue!40]  
     \centering
     \begin{minipage}[t]{.60\textwidth}

     \resizebox{\columnwidth}{!}{%
          \begin{tikzpicture}[node distance=0.6cm]
          \node[decisions](rootnode){ $\mathbf{x}[1]\leq 1?$ };
          \node[decisions,below of =rootnode,yshift=-0.3cm,xshift=-1.5cm](point1){$\mathbf{x}[2]\leq 4?$};
          \node[decisions,below of =rootnode,yshift=-0.3cm,xshift=1.5cm](point2){$\mathbf{x}[3]\leq 3?$};
          \node[decisions,below of =point1,yshift=-0.3cm,xshift=-1.0cm](point3){$\mathbf{x}[2]\leq 2?$};
          \node[results,below of =point1,yshift=-0.3cm,xshift=1.0cm](leaf){$v_4$};
          \node[results,below of =point2,yshift=-0.3cm,xshift=-1.0cm](leaf1){$v_5$};
          \node[results,below of =point2,yshift=-0.3cm,xshift=1.0cm](leaf2){$v_6$};
          \node[results,below of =point3,yshift=-0.3cm,xshift=-1.0cm](leaf3){$v_1$};
          \node[decisions,below of =point3,yshift=-0.3cm,xshift=1.0cm](point4){$\mathbf{x}[4]\leq 5?$};
          \node[results,below of =point4,yshift=-0.3cm,xshift=-1.0cm](leaf4){$v_2$};
          \node[results,below of =point4,yshift=-0.3cm,xshift=1.0cm](leaf5){$v_3$};
          \draw [arrow] (rootnode) -- node [left,font=\small] {True} (point1);
          \draw [arrow] (rootnode) -- node [right,font=\small] {False} (point2);
          \draw [arrow] (point1) -- node [left,font=\small] {True} (point3);
          \draw [arrow] (point1) -- node [right,font=\small] {False} (leaf);
          \draw [arrow] (point2) -- node [left,font=\small] {True} (leaf1);
          \draw [arrow] (point2) -- node [right,font=\small] {False} (leaf2);
          \draw [arrow] (point3) -- node [left,font=\small] {True} (leaf3);
          \draw [arrow] (point3) -- node [right,font=\small] {False} (point4);
          \draw [arrow] (point4) -- node [left,font=\small] {True} (leaf4);
          \draw [arrow] (point4) -- node [right,font=\small] {False} (leaf5);
          \end{tikzpicture}
     }

     \end{minipage}
     \caption{A Binary Decision Tree} 
     \label{Fig.main1} 
\end{figure}
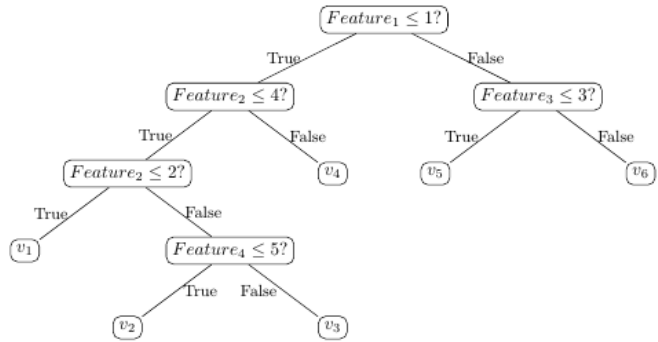

It is the results rather than the execution order of the Boolean tests that determine the prediction value of the decision trees for a specific input.
If we have known all the results of the tests, we want to directly guide the input from the root to the exit leaf.

The template vector of a leaf node  is defined by the following steps:
\begin{enumerate}
     \item each element corresponds to a distinct internal node;
     \item if the leaf is in the right branch of some ancestor node, the element attached to this ancestor is set to $1$;
     \item if the leaf is in the left branch of some ancestor node, the element attached to this ancestor is set to $-1$;
     \item if an internal node  is not a ancestor of the leaf, the element attached to this ancestor is set to $0$.
\end{enumerate}
For example, the template vector of the leaf node $v_3$ is $(-1, -1, 0, +1, +1)$ in the example \ref{Fig.main1}.
The template vectors are distinct from each other because each path is unique in decision trees.

If the input takes all the tests, it is supposed to follow the `maximum likelihood path'. 
The key step is to compare the similarities between the result vector $\mathbf{h}\in\{1, -1\}^{L-1}$ with the template vectors in $\{1, 0, -1\}^{L-1}$.  

We modify the cosine similarity into so-called logical similarity defined as 
\begin{equation}\label{zhang_similarity}
LS(\mathbf{k}, \mathbf{h})=\frac{\mathbf{k}\cdot\mathbf{h}}{\|\mathbf{k}\|_2^2},
\end{equation}
where the positive integer $\|k\|_2^2$ is the squared $\ell_2$ norm of template vector $\mathbf{k}\in\{1, 0, -1\}^{L-1}$.
Our similarity\eqref{zhang_similarity} is asymmetric and bounded by $(-1, 1]$.

This scheme would enjoy the following properties:
\begin{itemize}
  \item Each leaf node must be encoded as a vector of fixed length uniquely.
  \item If instance $\mathbf{x}$ can reach the leaf node $i$ with template vector $\mathbf{k}$, 
     then the similarity $LS(\mathbf{k}, h(\mathbf{x}))$ is equal to $1$. 
  \item If instance $\mathbf{x}$ can reach the leaf node $i$ with template vector $\mathbf{k}$, 
     then the similarity $LS(\mathbf{k}, h(\mathbf{x}))$ is the maximum of all possible paths.
\end{itemize}

For a binary decision tree, the template matrix $\mathrm{B}$ consists of all the template vectors as its rows.
We compute the similarity function \eqref{zhang_similarity} on a set of normalized template vectors simultaneously, 
packed together into a matrix $\bar{\mathrm{B}}$ as shown below
\begin{equation}\label{comparision}
\begin{split}
\bar{\mathrm{B}}\, h(\mathbf{x})  &= (\frac{B_1 h(\mathbf{x})}{\|B_1\|_2^2}, \cdots, \frac{B_L h(\mathbf{x})}{\|B_L\|_2^2})^T \\
                 &= (LS(B_1, h), \cdots, LS(B_L, h))^T
\end{split}
\end{equation}
where the template vector $B_i\in \{1, 0, -1\}^{L-1}$ is the $i$th row of matrix $\mathrm{B}$; 
the positive integer $\|B_i\|_2^2$ is the squared $\ell_2$ norm of the vector $B_i$ for $i=1,2,\cdots, L$;
the operator $LS(\cdot, \cdot)$ is the logical similarity defined in \eqref{zhang_similarity}.

The decision trees take the mode of the ground truth within each leaf node as the output prediction.
In another word, it is the leaf node that determines the prediction rather than the internal nodes.

We combine \eqref{result} and \eqref{comparision} and 
perform an interleaved traversal of binary trees in the means of matrix computation by
\begin{equation}\label{tree}
\begin{split}
     \mathbf{h} & = \operatorname{sgn}(\mathrm{S}\mathbf{x}-\mathbf{t}), \\
     \mathbf{p} & = \bar{\mathrm{B}}\mathbf{h}\in\mathbb{R}^L, \\
    v[i] & =v[\arg\max{(\mathbf{p})}],
\end{split}
\end{equation}
where the function $\arg\max(\mathbf{p})$ returns the index of  maximums of $\mathbf{p}$; the index $i$ is equal to $\arg\max(\mathbf{p})$; 
the output $v[i]$ is the value of the $i$-th leaf node for $i=1,2,\cdots, L$.
We call the new formulation of binary decision trees \emph{logical decision machines}.

Here we do not specify the data type or content form on the value of each leaf.
For example, the element $v[i]$ can linear function.
Usually it is numerical for regression and categorical/discrete for classification\footnotemark.
\footnotetext{See more differences of classification trees and regression trees in \cite{loh2011classification}.}
And we can embed the class labels into row vectors to make a value matrix $V$, 
where each column is the class label embedding vector associated with some leaf for classification.

Suppose the input variable $\mathbf{x}$ has 4 numerical features and 
the trained decision tree is as shown in the figure \ref{Fig.main1},
we will take a specific example to show how logical decision machine\eqref{tree} works.

The threshold vector is $\mathbf{t}=(1,4,3,2,5)^T$.
The value vector may be  categorical or numerical,
such as $\mathbf{v}=(v_1, v_2, v_3, v_4, v_5, v_6)^T$.
The selection matrix  and  template matrix  of the tree \ref{Fig.main1} are 
$$\mathrm{S}=\begin{pmatrix}
1 & 0 & 0 & 0\\
0 & 1 & 0 & 0\\
0 & 0 & 1 & 0\\
0 & 1 & 0 & 0\\
0 & 0 & 0 & 1
\end{pmatrix},\quad
\mathrm{B}=\begin{pmatrix}
-1 & -1 & 0 & -1 & 0\\
-1 & -1 & 0 & +1 & -1\\
-1 & -1 & 0 & +1 & +1\\
-1 & +1 & 0 & 0 & 0\\
+1 & 0 & -1 & 0 & 0\\
+1 & 0 & +1 & 0 & 0\\
\end{pmatrix}.$$

Suppose the input instance $\mathbf{x}$ is $(2,1,2,2)^T$, 
its prediction is $v_5$ according to the figure \ref{Fig.main1}.
The computation procedure of logical decision tree is as shown below
$$\mathrm{S}\mathbf{x}-\mathbf{t}=\begin{pmatrix} 1\\ -3\\ -1\\ -1\\ -3 \end{pmatrix},\quad
\mathbf{h}=\operatorname{sgn}(\mathrm{S}\mathbf{x}-\mathbf{t})=\begin{pmatrix} 1\\ -1\\ -1\\ -1\\ -1 \end{pmatrix},\quad
\mathbf{p}=\bar{\mathrm{B}} \mathbf{h}=\begin{pmatrix}\frac{1}{3}\\ 0\\ -\frac{1}{2}\\ 0\\ 1\\0 \end{pmatrix},$$
$$i=\arg\max(\mathbf{p})=5, \mathbf{v}[i]=v_5.$$ 
In summary, we obtain $T(\mathbf{x})=v_5$.
 
\subsection{Provable guarantee}

The template matrix $\mathrm{B}$ in \eqref{tree} is related with the structure of decision tree.
  
The sibling leaves share the same parent and ancestor nodes 
thus they share the same tests and their unique different result is returned by their parent node.
From the computation perspective, the difference of sibling template vectors is parallel to certain one-hot vector.

Each leaf node is attached to a row vector in the template matrix $\mathrm{B}$.
According to definition of template vector, each internal node is attached to a column vector in the template matrix $\mathrm{B}$,
where its element is attached to a leaf node.
The root node will test all samples so that its corresponding results are not zero in all template vectors.
In another word, there is a column vector without any 0 in $\mathrm{B}$. 
For example, the first column of $\mathrm{B}$ of the decision tree shown in \ref{Fig.main1} is associated with the test of the root,
which means the test is true for the first four leaves and false for the last two ones.
Thus, it implies that the first four leaves belongs to left subtree of the root node and
the last two ones belong to the right subtree of the root node.
Every test will produce two results, so each column must have at least one $1$ and $-1$.
In another word, each test must be involved with at least two decision paths.

\begin{theorem}\label{rank}
     The template matrix $\mathrm{B}$ is of full column-rank.
\end{theorem}

\begin{proof} 
    It is equivalent to prove that the column vectors of $\mathrm{B}$ are linearly independent.

    For simplicity, we call the column vectors of $\mathrm{B}$ in the right subtree as right vectors 
    and the column vectors of $\mathrm{B}$ in the left subtree as left vectors.

    The right vectors are independent of the ancestors of the left leaves, 
    so the corresponding elements of left nodes in the right vectors are zero.
    Thus, those zeroes can lead to the linear independence of the right vectors and the root vector.
    Similarly, we can prove the linear independence of the left vectors and root vector.

    We apply  the above procedure to all the subtrees and complete the proof.
\end{proof}

It is not difficult to recover the hierarchical structure of certain binary  tree
if we have known its template matrix. 
The template matrix is agnostic with respect to the test at the internal branch and the prediction at the leaf node.

Additionally, we can find the subtree is associated with the  sub-matrix of the template matrix. 
For example, we can find  the template matrix of the left subtree in \ref{Fig.main1} as following 
$$\begin{pmatrix}
      -1  & -1 & 0 \\
      -1  & +1 & -1\\
      -1  & +1 & +1\\
      +1  & 0  & 0 \\
\end{pmatrix}.$$

Now we prove the correctness of the algorithms \eqref{tree} and 
build a connection between decision trees and error-correcting output codes.
\begin{theorem}\label{correctness}
     The conversion algorithms \eqref{tree} is correct.
\end{theorem}
\begin{proof} 

We want to prove that $LS(b_i, h(\mathbf{x}))=1$ if and only if the instance $\mathbf{x}$ can reach the leaf $i$
where the vector $b_i$ is the $i$th template vector of a decision tree for $i\in \{1, \cdots, L\}$ 
and $h(\mathbf{x})$ is the result vector for an input instance $\mathbf{x}$, 

If the sample $\mathbf{x}$ reach the leaf node $i$, it is easy to verify that $LS(b_i, h(\mathbf{x}))=\max_{i\in\{1,2,\cdots, L\}}LS(b_i, h(\mathbf{x}))=1$.

If $LS(b_i, h(\mathbf{x}))=1$, we can obtain $\left<b_i, h(\mathbf{x}) - b_i\right>=0$  by the definition \eqref{zhang_similarity} 
so $h(\mathbf{x})$ is equal to $b_i$ for every non-zero element in $b_i$, 
which means the instance $\mathbf{x}$ can satisfy the conditions of the $i$th leaf's ancestor nodes.
Becasue the exit leaf node is unique for every sample, so the the maxima in \eqref{comparision} is unique.
Thus, the maxima in \eqref{comparision} is sufficient to identify the exit leaf. 
And we complete the proof.
\end{proof}

\subsection{Consistent interpretation}
The logical decision machines are consistent with some machine learning models.

We call the proof \ref{correctness} the \emph{template matching interpretation} of decision trees
because it is to select the prediction value based on similarity between  template vectors  and the result vector of instances.
We can obtain the convex-concave upper bound on empirical loss proved in \cite{norouzi2015efficient}
and use the non-greedy methods therein to train the logical decision machines. 

With the test vectors, we can approximate the prediction value of the $j$th leaf node 
as a kernel based on random forests estimate in \cite{kernel2016} 
\begin{equation}
\ell_j.val
=\sum_{(\mathbf{x}_i, y_i)\in \mathcal{D}}\frac{y_i\delta(\|\mathbf{h}, \mathbf{h}_i\|_{\mathrm{B}})}{N_j}
\end{equation}
where the function $\delta(z)$ is equal to $1$ if $z=0$ and $0$ otherwise;
$\mathbf{h}$ and $\mathbf{h}_i$ are the hidden vectors of new instance $\mathbf{x}$ and $\mathbf{x}_i$ respectively; 
the number $N_j$ is the total sum of samples falling in the $j$th leaf node;
the matrix $\mathrm{B}$ is the template matrix;
the notation $\|\mathbf{h}, \mathbf{k}\|_{\mathrm{B}}$ is a distance defined as $\|\mathrm{B}(\mathbf{h}-\mathbf{k})\|_2$.

The algorithm \ref{tree} is also similar to the error correcting output codes (ECOC) 
such as \cite{dietterich1994solving, Kong95error-correctingoutput, allwein2001reducing}, 
which is a powerful tool to deal with multi-class categorization problems.
The basic idea of ternary ECOC is to associate each class $r$ with a row of a coding matrix $\mathrm{M} \in \{-1, 0, +1\}^{k \times \ell}$.
The encoding stage of ECOC is described  in the algorithms\eqref{ECOC}.

\begin{algorithm}[h]\label{ECOC}
	\caption{Ternary Error Correcting Output Codes}
	\begin{algorithmic}[1]
		
		\Statex \textbf{Input}:
		\begin{itemize}
			\item  the test example $\mathbf{x}$.
			\item  the coding matrix $\mathrm{M} \in \{-1, 0, +1\}^{k \times \ell}$ for class $1,2, \cdots, k$.
			\item  the pre-trained binary classifiers $f=\{f_s\mid s=1, 2, \cdots, \ell\}$ for each column of $\mathrm{M}$.

		\end{itemize}
          \Statex \textbf{Output}: the predicted label
		\Procedure{Encoder}{$\mathbf{x}, f$} 
          \State Compute the binary predictions of $\mathbf{x}$: $\mathbf{p}=(f_1(\mathbf{x}), \dots, f_{\ell}(\mathbf{x}))$
		\For{ row $\mathbf{r}\in \mathrm{M}$}  \Comment{iterate over the rows}
		\State $d_r \leftarrow \mathrm{dis}(\mathbf{r}, \mathbf{p})$  
		\EndFor 
		\State $j\leftarrow$ index of the minimum in $\{d_1, d_2, \cdots, d_{\ell}\}$
		\State return $j$
		
		\EndProcedure

	\end{algorithmic}
\end{algorithm}

The bit-vector corresponds to the error-correcting code $h$ and 
the bit-vector matrix $\mathrm{B}$ corresponds to the error-correcting code matrix $\mathrm{M}$.
All internal nodes are associated with binary test, 
which are parameterized by $\mathrm{S}$ and $\mathbf{t}$.
The main difference between \ref{ECOC} and \eqref{tree} is that the row in the code matrix $M$ corresponds to a specific class
while the row in the template matrix $\mathrm{B}$ corresponds to a specific leaf. 
In one hand, we can consider our method as another coding strategy of  ternary error-correcting output codes.
In another hand, the coding strategy such as \cite{escalerasergio2010on} will help to find new induction methods of decision tree.
Their main difference is that each class is attached with a single code word in ECOC 
while each class may be associated with multiple leaf nodes in decision trees.

\section{Analytical Decision Machines}\label{sec3}

To elucidate the probability concentration on the exit leaf, we have reformulated the process \eqref{tree} into a more analytical framework, 
delineating three key equations that capture the essence of our approach.
\begin{equation}\label{def:LPOM}
\begin{split}
\mathbf{h} &= \operatorname{sgn}(\mathrm{S}\mathbf{x}-\mathbf{t}),\\
\mathbf{p} &= \arg\max_{\mathbf{d}\in\Delta}\left<\mathbf{d}, \mathrm{\bar{B}}\mathbf{h}\right>,\\
T(\mathbf{x}) &=  \sum_{i=1}^{L}p[i]v[i],
\end{split}
\end{equation}
where the notation $\Delta$ is the probability simplex defined as $\Delta=\{\mathbf{d}\mid \mathbf{d}\geq 0, \mathbf{d}^T\vec{1}=1\}\subset \mathbb{R}^L$;
the optima $\mathbf{p}$ is in $\{0, 1\}^L$ where its only non-zero element is in the position of maximum in $\mathbf{h}\in \mathbb{R}^L$; 
the integer $L$ is the total number of leaves.
Initially, the hidden decision $\mathbf{h}$ for an input instance $\mathbf{x}$ is calculated through the feature selection matrix 
$\mathrm{S}$ and the threshold vector $\mathbf{t}$. 
Subsequently, the most probable exit leaf is determined by maximizing the dot product within the probability simplex $\Delta$.
Finally, the ultimate prediction $T(\mathbf{x})$ is derived as a weighted sum of the values associated with each leaf, 
reflecting their respective probabilities.
Clearly, the logical decision machine \eqref{tree} is of hard attention mechanism\cite{xu2015show}.

Because the only maximum value is equal to $1$ in the similarities defined by \eqref{comparision}, 
we re-express \eqref{tree} in the following way
\begin{equation}\label{delta}
     \begin{split}
          \mathbf{h}    &=\mathrm{\bar{B}}\operatorname{sgn}(\mathrm{S}\mathbf{x}-\mathbf{t}),\\
          T(\mathbf{x}) & = \sum_{i=1}^{L}\delta(1-\mathbf{h}[i])\mathbf{v}[i],
     \end{split}
\end{equation}
where the function $\delta(z)$ is equal to $1$ if $z= 0$ and $0$ otherwise.
And $\delta(\cdot)$ is activated to identify the exit leaf.
Obviously, both $\mathrm{\bar{B}}$ and $\mathrm{S}$ are quite sparse matrices, 
which may help to understand why sparse or quantized neural networks work.
However, the shapes of $\mathrm{\bar{B}}$ and $\mathrm{S}$ may restrict the design space of logical decision machines.
Additionally, the operators $\operatorname{sgn}(\cdot)$, $\delta(\cdot)$ and $\arg\max(\cdot)$ are non-differentiable, 
which isolates logical decision machines from  deep neural networks.

The additive tree such as \cite{breiman2001random, chen2015xgboost} is in the form of 
\begin{equation}\nonumber
F(\mathbf{x})=\sum_{i=1}^N w_i T_i(\mathbf{x})
\end{equation}
where $w_i\in\mathbb{R}$ and the tree $T_i$ is the instances of \eqref{delta} for $i=1, 2,\cdots, N$.
It is simple to evaluate \eqref{forest} in parallel as the same as the random forest \cite{breiman2001random}.
For example, suppose there are two instances $T(\mathbf{x})$ and $\tilde{T}(\mathbf{x})$ of \ref{delta} as following 
$$\bar{\mathbf{h}} = \bar{\mathrm{B}} \operatorname{sgn}(\bar{\mathrm{S}}\mathbf{x} - \bar{\mathbf{t}}),\, 
{T}_1(\mathbf{x}) = \sum_{i=1}^{L_{1}}\delta(1-\bar{\mathbf{h}}[i])\bar{\mathbf{v}}[i], $$
$$\tilde{\mathbf{h}} = \tilde{\mathrm{B}}\operatorname{sgn}(\tilde{\mathrm{S}}\mathbf{x} -\tilde{\mathbf{t}}),\, 
{T}_2(\mathbf{x}) = \sum_{i=1}^{L_{2}}\delta(1-\tilde{\mathbf{h}}[i]) \tilde{\mathbf{v}}[i], $$
we can evaluate them in parallel and obtain the sum of trees $w_1\bar{T}+w_2\tilde{T}$.
In fact, the sum of the decision trees is yet another instance of \eqref{delta} as proven below:
\begin{equation}\label{forest}
     \begin{split}
          \mathbf{h} &= \mathrm{B}\operatorname{sgn}(\mathrm{S}\mathbf{x} - \mathbf{t}), \\
          T(\mathbf{x}) &= \sum_{i=1}^{L_1+L_2}\delta(1-\mathbf{h}[i])  \mathbf{v}[i], 
     \end{split}
\end{equation}
where 
$\mathrm{S}=\begin{pmatrix}
    \bar{\mathrm{S}} \\ \tilde{\mathrm{S}}
\end{pmatrix}, 
\mathbf{t}=\begin{pmatrix}
     \bar{\mathbf{t}} \\ \tilde{\mathbf{t}}
\end{pmatrix},
\mathrm{B}=\begin{pmatrix}
     \bar{\mathrm{B}} & 0 \\
     0 &\tilde{\mathrm{B}}
\end{pmatrix}, 
\mathbf{v}=\begin{pmatrix}
     w_1 \bar{\mathbf{v}} \\
     w_2 \tilde{\mathbf{v}}
\end{pmatrix}$.
Here we put the multipliers in the value vectors; 
the function $\delta(\cdot)$ is activated twice for each input example $\mathbf{x}$.
By this method, we remove the restriction on the shapes of the matrices $\bar{\mathrm{B}}$ and $\mathrm{S}$ in \eqref{tree}.
We are closer to the key-value attention \cite{mino-etal-2017-key, vaswani2017attention}. 

In the rest of this section, 
we will construct analytical decision machines to approximate or remove some operators in logical decision machines
and take advantages of attention mechanisms to expand the design space of decision machines.

\subsection{Construction}

We begin from  presenting a simple relaxation of decision machines\eqref{delta} and, 
then, we introduce three crucial refinements to the algorithm \eqref{tree}.

We can relax $\operatorname{sgn}(\cdot)$ and $\delta(\cdot)$ in \eqref{delta} by standard approaches without any modifications of the rest.

The first refinement is to remove the $\operatorname{sgn}$ function defined in \ref{signum}. 
The exit leaf is identified by the position of the leftmost bit set to $1$ in the product of the false nodes' bit-vectors,
which are assigned to the internal nodes to represents false leaves as $1$ and true leaves as $0$ in $\text{QuickScorer}$ \cite{lucchese2015quickscorer}.
Similarly, it is able to identify the exit leaf by the position of the rightmost bit set to $1$ in the product of the true nodes' bit-vectors
if the false nodes and true nodes are represented as $0$ and $1$ in the bitvectors respectively.
The $\operatorname{sgn}(\cdot)$ operator in \eqref{def:LPOM} is removed by combining both procedures
with well-designed matrices  $\mathrm{F}$  as described below
\begin{equation}\label{defK}
     \begin{split}
          \mathbf{h}    &= \mathrm{F}\operatorname{ReLU}(\mathrm{W}\mathbf{x}-\mathbf{b}),\\
          T(\mathbf{x}) & = \mathbf{v}[\arg\max(\mathbf{h})],
     \end{split}
\end{equation}
where the operator $\operatorname{ReLU}(\mathrm{W}\mathbf{x}-\mathbf{b})$ is defined as $\max(0, \mathrm{W}\mathbf{x}-b)$;
$\arg\max(\cdot)$ returns the smallest index of the maximum value in the input vector $\mathbf{h}$.

The second refinement turns the algorithm \eqref{defK} into a soft decision tree, 
where instances are directed to the exit leaf with some confidence calculated by a scoring function.
We apply differentiable $\operatorname{softmax}(\cdot)$ to smoothen distribution over the leaf values in \eqref{defK}:
\begin{equation}\label{GT2}
 T(\mathbf{x}) =\sum_{i=1}^{L}\alpha_i \mathbf{v}_i,
\end{equation}
where $(\alpha,\cdots,\alpha_L)^T=\operatorname{softmax}(\frac{\mathrm{K}\, (\mathrm{S}\mathbf{x}-\mathbf{t})}{\tau})$; 
the optional $\tau$  is the temperature hyperparameter in $(0, 1)$.
There are diverse variants of $\operatorname{softmax}$ such as \cite{jang2016categorical, zhang2018weighted, Sigsoftmax1, laha2018controllable}.
Here we focus on the positively correlated tests and  define our alternative to the $\operatorname{softmax}$ operator in \eqref{GT2} by
\begin{equation}\nonumber
     \sum_{i=1}^L \frac{\alpha\, sim(K_i, q)_{+}+1-\alpha}{\alpha\,S+L(1-\alpha)}v_i,
\end{equation}
where the vector $q=Sx-t$; the real $sim(K_i, q)_{+}$ is the positive part of the $sim(K_i, q)$ defined by $\max(0, \left<K_i, q\right>)$; 
 $\mathrm{S}$ is $\sum_{i=1}^L sim(K_i, q)_{+}$; the optional $\alpha$ is a hyperparameter in $(0,1]$.
It is a special case of \cite{Sigsoftmax1}.

The third  refinement is aimed at realization of the soft decision tree\cite{irsoy2012soft}.
In \cite{irsoy2012soft}, each internal node directs instances to its children 
with probabilities given by a sigmoid gating function.
It is equivalent to assign a vector to each branch 
where left leaves are represented as the sigmoid function $\sigma(\cdot)$; 
right leaves are represented as $1-\sigma(\cdot)$; 
and other leaves are represented as $1$.
We can derive a matrix $\mathrm{M}$ from the template matrix $\mathrm{B}$ 
to replace $\bar{\mathrm{B}}$ in \eqref{tree} as shown below
\begin{equation}\label{defM}
     \mathrm{M}_{ij}  = 
     \begin{cases}
          \sigma(\mathrm{B}_{ij}(\mathbf{w}_j^T \mathbf{x}+\mathbf{b}_j)), &\text{if $\mathrm{B}_{ij}\neq 0$,}\\
          1, &\text{otherwise},
     \end{cases}
\end{equation}
where the function $\sigma(z)=\frac{1}{1+\exp(-z)}$;
the integer $\mathrm{B}_{ij}\in\{1, 0, -1\}$ is the $j$th element in the $i$th row of template matrix $\mathrm{B}$ in \eqref{comparision};
the weight $\mathbf{w}_j\in\mathbb{R}^n$ and the bias $b_j\in\mathbb{R}$ are the parameters associated with $j$th branch.
The probability $p[i]$ for each leaf $i$ is equal to $\prod_{j=1}^{L-1}\mathrm{M}_{ij}$, 
which is the product of all the non-zero elements in the $i$th row of $\mathrm{M}$.
With some arithmetic tricks, the output $y$ of the soft decision tree \cite{irsoy2012soft} is evaluated by
\begin{equation}\label{soft}
     \begin{split}
     \mathbf{p}  &= \exp(\ln(\mathrm{M})\vec{1}),\\
     y  &= \sum_{i=1}^L \mathbf{p}[i] \mathbf{v}[i],
     \end{split}
\end{equation}
where the matrix $M$ is defined in \eqref{defM}; 
the vector $\vec{1}$ is a vector of ones with the same row length of $\mathrm{M}$;
the vector $\mathbf{p}$ is the probability distribution of each leaf, 
i.e.,$\sum_{i=1}^L p[i]=1$.

Matrices in \eqref{tree}, \eqref{forest}, \eqref{defK} and \eqref{defM} contain diverse hierarchical information,
which makes it distinct from the nested logit model and hierarchical softmax \cite{pmlr-vR5-morin05a, hlm2008, hsc2018}.

\subsection{Attention Models}


Great success of the Transformer architecture proposed in \cite{vaswani2017attention}
benefits from integral combination of its components.
In this section, we will lend some support to the design choices of Transformer 
and borrow some ideas from them to expand the design space of decision machines.

Now we turn to the scaled dot-product attention in \cite{vaswani2017attention} defined by
$$Attention(\mathrm{Q}, \mathrm{K}, \mathrm{V})=\operatorname{softmax}(\frac{\mathrm{Q}\mathrm{K}^T}{\sqrt{d_k}})\mathrm{V},$$
where a set of queries are packed together into a matrix $\mathrm{Q}$ in order to compute the attention function on these queries simultaneously.
We can evaluate the attention output $\mathbf{y}_i$ for each query $\mathbf{q}_i$ separately as follows
\begin{equation}\label{dot-product}
     \mathbf{y}_i = \operatorname{softmax}(\frac{\mathbf{q}_i \mathrm{K}^T}{\sqrt{d_k}})\mathrm{V}=
     \sum_{j=1}^m \frac{sim(\mathbf{k}_j, \mathbf{q}_i)}{\sum_{n=1}^i sim(\mathbf{k}_n, \mathbf{q}_i)} \mathbf{v}_j,
\end{equation}
where the term $sim(\mathbf{k}_j, \mathbf{q}_i)$ is defined as 
$\exp(\left< \mathbf{k}_j, \mathbf{q}_i \right>)$
 to measure  the similarity between the query $\mathbf{q_i}$ and
the key $\mathbf{k_j}$; the integer $d_k$ is the dimension of the key $\mathbf{q_i}$.
Our algorithm \eqref{GT2} is almost identical to the scaled dot-product attention \eqref{dot-product}, 
except for the divided temperature of $\tau$.
It is a strong evidence  that we can derivate the attention mechanism from the decision tree.
We relaxed the binary decision tree in \eqref{GT2} from the perspective of model representation,
while the scaled dot-product attention is used to let sequence transduction models dispense with recurrence and convolutions in \cite{vaswani2017attention}.

As shown before, it is the key matrix $\mathrm{B}$ that determines the intrinsic hierarchical structure of \eqref{forest}.
We can mask the sub-structure of additive trees by matrix projection.
For example,
We can mask the second tree in \eqref{forest}:
\begin{equation}\nonumber
     \begin{split}
          h_{B} &= \mathrm{B}W^{B}\operatorname{sgn}(\mathrm{S}\mathbf{x} - \mathbf{t}), \\
          T_{B}(x) &= \sum_{i=1}^{L_1+L_2}\delta(1-h_{B}[i])  v[i], 
     \end{split}
\end{equation}
where the matrix $W^{B}=\begin{pmatrix}1_{L_1\times (L_1 - 1)} & 0 \\
     0 & 0_{L_2\times (L_2 - 1)}
\end{pmatrix}$; 
the rest are the same as in \eqref{forest}.
Here we projected $\mathrm{B}$ of \eqref{forest} to  $\begin{pmatrix}\bar{B}_1 & 0 \\
     0 & 0
\end{pmatrix}$.
This finding may provide some insights into multi-head attention,
which originated in \cite{vaswani2017attention} to capture the information from different representation subspaces at different positions 
\begin{equation}\label{MultiHead}
     \begin{split}
          \text{MultiHead}(Q, K, V )  & =  \text{Concat}(head_1, \dots, head_h)W^O \\
          \text{where $head_i$} & = Attention(QW_i^Q, KW_i^K, VW_i^V)
     \end{split}
\end{equation}
where the queries, keys and values are projected into different subspaces by $QW_i^Q$, $KW_i^K$ and $VW_i^V$, respectively.

We can borrow the ideas from sparse attention 
\cite{niculae2017regularized, vaswani2017attention, malaviya2018sparse, correia2019adaptively, gupta2021memoryefficient} 
or linear transformer \cite{tsai-etal-2019-transformer, kathRNN2020} to expand the design space of decision machines, 
which are used to reduce the quadratic complexity of scaled dot-product attention in \cite{vaswani2017attention}.
The first is to use the sparse alternatives to $\operatorname{softmax}$ 
such as \cite{martins2016softmax, niculae2017regularized, malaviya2018sparse, correia2019adaptively, gupta2021memoryefficient}. 
We prefer these sparse alternatives to $\operatorname{softmax}$ for the sake of more interpretability.
The second is to replace the direct computation of similarity between $K$ and $Q$ with a transformation 
through a non-linear kernel function in linear transformers such as \cite{tsai-etal-2019-transformer, kathRNN2020}.
Those advances in sparse attention and linear transformer permit an efficient acceleration of decision machines \eqref{GT2}.

\section{Discussion and Conclusion}\label{sec4}
In this paper, we have introduced Decision Machines (DMs), a novel extension of binary decision trees, 
which leverage matrix computation to enhance their computational efficiency and interpretability. 
By embedding the decision paths and test results into a vector space, 
DMs offer a more streamlined approach to model prediction. 
Our exploration into the integration of DMs with attention mechanisms has opened new avenues for leveraging deep learning toolkits to accelerate decision tree operations.

\bibliographystyle{plain}
\bibliography{tree}
\end{document}